\documentclass[runningheads]{llncs}

\usepackage{eccv}
\usepackage{eccvabbrv}

\usepackage{graphicx}
\usepackage{booktabs}
\usepackage{array}
\usepackage{placeins}
\usepackage{tikz}
\usetikzlibrary{positioning}
\usepackage[hidelinks]{hyperref}

\begin{document}

\title{Frozen High-Resolution Inference for Cross-City Object Detection: An AI City Challenge 2026 Study}
\titlerunning{Frozen High-Resolution Inference for Cross-City Detection}

\author{Jaeuk Kim}
\authorrunning{J. Kim}
\institute{NextITS, Seoul, Republic of Korea \\
\email{freak91uk@hnextits.com}}

\maketitle

\begin{abstract}
Cross-city object detection requires a detector trained in one city to generalize to an unlabeled target city. In AI City Challenge 2026 Track 6 we analyze archived configurations of a single RF-DETR-Large detector inside an air-gapped Training-as-a-Service platform whose server returns only an aggregate COCO-style AP over a hidden mixture of source- and target-city images. Frozen $1120\times1120$ inference of a checkpoint trained at $704\times704$ achieved the highest aggregate AP among the evaluated configurations ($0.3272 \rightarrow 0.3654$, $+0.0382$) without any parameter update, with the largest relative gain on small objects and the largest absolute gain on medium objects, at $2.53\times$ the input pixels. A warm-start 1120px fine-tuning recipe reached $0.3470$ while its in-domain validation AP rose ($0.767\rightarrow0.789$)---a caution that in-domain validation is an unreliable model-selection signal under aggregate-only cross-city feedback. Because that run's evaluation used a higher confidence threshold than the inference-only runs ($0.05$ vs.\ $0.01$), we treat its score as a descriptive archived outcome rather than a controlled verdict on fine-tuning. Gray-world normalization did not meaningfully change the frozen-1120 result, and a rectangular run was found by audit to have used an unintended portrait orientation. We release verbatim platform commands, configuration snapshots and an explicit evidence boundary for every claim. Each configuration was submitted once and the best was selected on the hidden server, so these are exploratory, audited findings about the aggregate mixture; they do not establish target-city-specific improvement.

\end{abstract}

\section{Introduction}

AI City Challenge 2026 Track 6 (Cross-City Object Detection, on the Milestone Project Hafnia platform)~\cite{aicity2026track6,hafnia_crosscity_dataset,tang2025aicity} evaluates detectors under a single-source domain-generalization protocol: models are trained on one source city and evaluated on a hidden benchmark that mixes source-city images with images from a distinct target city, over ten fine-grained traffic classes (nine vehicle subtypes plus \emph{Person}). The target city's images are unlabeled and unavailable at training time, and the challenge server returns a single \textbf{aggregate} COCO-style AP over the mixed benchmark; it does not separate the source- and target-city halves. This aggregate-only feedback, together with a limited number of server submissions, shapes how our findings must be read: because no per-domain metric is available, no result reported here can establish that an intervention improved target-city detection specifically. Every reported score characterizes the mixed benchmark. We keep the challenge's cross-city framing because that is the task the benchmark is built for, but we make no target-domain claim anywhere in this paper.

We participated with a single RF-DETR-Large detector (DINOv2 ViT-S backbone) and asked a narrow, practical question: \textbf{under a restricted single-source cross-city challenge with no target validation signal and a limited submission budget, how did practical inference- and training-time interventions affect the aggregate hidden-benchmark performance of a fixed RF-DETR model family?} We evaluated three intended interventions against a 704 baseline, and report one further rectangular configuration that the run-record audit later showed to have run in an unintended orientation; each configuration could be assessed only through a limited number of server submissions.

Our main empirical observation is that, among the configurations we evaluated, the highest aggregate score was obtained without any additional parameter update: \textbf{frozen $1120\times1120$ inference of a 704-trained checkpoint} raised the aggregate AP from $0.3272$ to $0.3654$ ($+0.0382$), driven by gains on medium and small objects. An alternative strategy that additionally adapts the model parameters to the higher training resolution---1120px warm-start fine-tuning---reached a lower aggregate score ($0.3470$) even though its in-domain validation stayed high---an outcome we report descriptively, since a run-record audit showed its evaluation settings were not uniform with the others (Section~\ref{sec:threshold}). Two inference-only interventions (rectangular resizing and gray-world normalization) did not improve over frozen 1120 inference.

We make three contributions:
\begin{enumerate}
\item \textbf{An audited, reproducible challenge study} under the restricted TaaS setting: verbatim platform commands for every reported run, a post-challenge run-record audit that surfaced non-uniform evaluation settings (Section~\ref{sec:threshold}), and an explicit evidence boundary for each claim (Table~\ref{tab:boundary}).
\item \textbf{Two empirical observations, each bounded.} (a) Among the configurations we could submit, frozen $1120\times1120$ inference achieved the highest aggregate hidden-benchmark AP with \emph{zero} parameter updates---largest relative gain on small objects, largest absolute gain on medium objects. (b) The one warm-start fine-tuning run we evaluated raised in-domain validation AP while its aggregate benchmark AP did not rise---a caution that \textbf{in-domain validation is an unreliable model-selection signal under aggregate-only cross-city feedback}.
\item \textbf{Honest negatives and scope.} The rectangular run used an unintended orientation rather than a designed ablation, gray-world normalization was negligible, and---because the metric is aggregate-only---no result can be attributed to the target domain.
\end{enumerate}
The contribution is not a new resizing operator. It is a checkpoint-shared, run-audited study comparing a parameter-preserving resolution change with an additional optimization recipe under restricted target access, aggregate-only feedback, and a limited submission budget---a statement about the evaluated configurations, the specific fine-tuning recipe, and this aggregate hidden benchmark, not about inference and fine-tuning in general.

\section{Related Work}

\noindent\textbf{Domain generalization and single-source DG for detection.} Domain generalization (DG) seeks models that transfer to unseen target distributions without target data, in contrast to domain adaptation, which assumes access to target samples~\cite{zhou2022dgsurvey}. For detection, most progress has followed the adaptation route, beginning with Domain Adaptive Faster R-CNN and its Cityscapes$\rightarrow$Foggy-Cityscapes benchmark~\cite{chen2018dafasterrcnn}. The harder, more realistic setting is single-domain generalized detection (Single-DGOD), formalized on urban cross-weather/cross-city data via cyclic-disentangled self-distillation~\cite{wu2022singledgod}. Recent Single-DGOD methods lean on trained interventions: vision-language semantic augmentation~\cite{vidit2023clipthegap}, source diversification with detection alignment~\cite{danish2024divalign}, and explicit debiasing of source-domain overfitting~\cite{liu2024unbiasedfasterrcnn}. Rather than evaluating a dedicated Single-DGOD method, we compare a simple warm-start high-resolution fine-tuning recipe with frozen resolution scaling on the same RF-DETR model family.

\noindent\textbf{DETR and RF-DETR.} DETR reframed detection as set prediction with a transformer encoder-decoder and bipartite matching, removing anchors and NMS~\cite{carion2020detr}. Deformable DETR added multi-scale deformable attention for faster convergence and scale handling~\cite{zhu2021deformabledetr}, and DINO's contrastive denoising and look-forward-twice recipe made DETRs both competitive and convergent~\cite{zhang2023dino}. RT-DETR then delivered a real-time, NMS-free end-to-end detector whose speed is tunable without retraining~\cite{zhao2024rtdetr}. RF-DETR, our detector, pairs this lineage with a neural-architecture-searched design and a windowed DINOv2 backbone~\cite{robinson2026rfdetr}; the self-supervised DINOv2 features are known for strong cross-domain robustness~\cite{oquab2024dinov2}.

\noindent\textbf{High-resolution inference and train-test resolution.} Scale is handled internally by feature pyramids~\cite{lin2017fpn}, yet detectors remain far from scale-invariant: train/test object-size mismatch measurably degrades accuracy~\cite{singh2018snip}, and input resolution trades directly against accuracy, especially for small objects~\cite{huang2017speedaccuracy}. Training-free high-resolution inference has also been studied through image slicing, as in SAHI~\cite{akyon2022sahi}, while QueryDet selectively activates high-resolution features to reduce the cost of small-object detection~\cite{yang2022querydet}. Prior work in image classification studied train--test resolution mismatch and resolution-dependent adaptation~\cite{touvron2019fixres}; our setting differs in that we evaluate a \emph{frozen object detector} on a hidden cross-city benchmark without target data or parameter updates. The RF-DETR implementation interpolates positional embeddings to accommodate the larger input grid, enabling evaluation of the frozen checkpoint at $1120\times1120$; related ViT research has also studied flexibility across patch and input configurations~\cite{beyer2023flexivit}. Our intervention differs from slicing and query-based methods in that it applies only a global input-resolution change to a frozen checkpoint---without slicing, regional refinement, architecture modification, or target adaptation. Unlike test-time adaptation, which updates weights on target data~\cite{wang2021tent}, it touches no parameters.

\noindent\textbf{Self-training and UDA for detection.} Cross-domain detection is often addressed with EMA teacher-student self-training~\cite{tarvainen2017mean}, confidence-thresholded pseudo-labeling~\cite{liu2021unbiased}, mean-teacher variants~\cite{deng2021unbiased}, adversarial adaptive teachers~\cite{li2022cross}, and probabilistic pseudo-box modeling~\cite{chen2022probabilistic}. All require target-domain images at training time, and confirmation bias---overfitting to one's own wrong pseudo-labels~\cite{arazo2020pseudo}---is a known failure; in the Track 6 protocol the target/benchmark images are not part of the public training workflow. Our main intervention instead changes only the inference resolution and requires neither target images nor parameter updates.

\noindent\textbf{Box fusion and ensembling.} Aggregating multi-scale or multi-model outputs relies on post-processing: greedy NMS~\cite{neubeck2006efficientnms}, its score-decay~\cite{bodla2017softnms}, distance-aware~\cite{zheng2020diou}, and learned~\cite{hosang2017learningnms} variants, and box-merging via Weighted Box Fusion~\cite{solovyev2021wbf}, the recipe behind challenge-winning ensembles~\cite{gu2020vipriors}. We cite WBF as relevant background but do not include it in our final evaluated configuration set; we evaluate gray-world normalization~\cite{buchsbaum1980grayworld} and report it as a negative result in our setting.

\section{Challenge Protocol and Experimental Setup}

\subsection{Dataset and challenge protocol}
Track 6 provides fine-grained traffic imagery for a single source city and evaluates on a hidden benchmark that mixes source-city images with images from a distinct target city~\cite{aicity2026track6,hafnia_crosscity_dataset}. Ten classes are scored: Car, Van, Pickup Truck, Single-unit Truck, Combo/Articulated Truck, Heavy-duty Truck, Trailer, Motorcycle, Bicycle, and \emph{Person}. The training and validation data contain approximately 13{,}000 frames and 150{,}000 annotated object instances; the hidden benchmark is reported to contain a comparable number of frames and instances, although exact per-split and domain-specific statistics are not disclosed. Training, inference, and benchmarking run inside the air-gapped Hafnia Training-as-a-Service (TaaS) platform; the full training and benchmarking datasets are not downloadable (only a small sample is provided for local development), external training data is prohibited, and only one experiment runs at a time. The public Track 6 workflow provides training data only inside a managed training job and evaluates a submitted inference package against the hidden benchmark; benchmark images are never exposed to participants.

\subsection{Backbone and platform}
RF-DETR-Large (rfdetr 1.8.1; DINOv2 ViT-S windowed backbone). The base model was trained for 80 epochs at $704\times704$ with a cosine LR schedule and a domain-oriented photometric/geometry augmentation preset (\texttt{dg\_crosscity\_v2}). The allowed pretrained RF-DETR (DINOv2) weights were bundled into the trainer package; no external task-specific data was used at any stage. The best checkpoint was selected by validation AP (the higher of EMA and regular weights); for the base model this was the EMA checkpoint, which we denote the ``704 EMA checkpoint'' and use for all inference-only configurations (L0--L2 and L4) and as the warm-start for L3. Table~\ref{tab:train} lists the training configuration; the random seed and per-operation augmentation probabilities are documented in the supplementary material.

\begin{table}[t]
\centering
\caption{Base training configuration.}
\label{tab:train}
\begin{tabular}{ll}
\toprule
Item & Setting \\
\midrule
Model & RF-DETR-Large (rfdetr 1.8.1) \\
Backbone & DINOv2 ViT-S, windowed \\
Train resolution & $704\times704$ \\
Epochs & 80 \\
Effective batch & 16 (batch 1 $\times$ grad-accum 16) \\
Optimizer & AdamW \\
Learning rate & 7e-5 (decoder); 1e-4 (encoder) \\
Scheduler & cosine, 1-epoch warmup \\
Precision & mixed (AMP) \\
num\_select & 300 \\
Augmentation & dg\_crosscity\_v2 \\
Weight decay & 1e-4 \\
EMA decay & 0.993 \\
Checkpoint selection & best validation AP (EMA/regular) \\
Hardware & Hafnia T4-class GPU \\
External task-specific data & None \\
\bottomrule
\end{tabular}
\end{table}

\subsection{Evaluation and model-selection protocol}
The challenge server reports a single COCO-style AP (IoU $0.50{:}0.05{:}0.95$, maxDets $=100$ per image) over a hidden mixture of source-city and target-city images. It does not provide domain-separated metrics, per-image evaluation results, benchmark images, or target annotations. During the live competition the server displayed a 50\% subset; the numbers here are the full-test scores. Each configuration was submitted once and evaluated only by the server; there is no local target ground truth. We compared a small, limited set of configurations and kept the best-scoring one---so the challenge server also played a limited \emph{model-selection} role, which we account for in the limitations.

\subsection{Run-record audit and comparison caveats}
\label{sec:threshold}
An audit of the archived platform run records, carried out after the challenge closed, identified two settings that were not uniform across all reported configurations: the evaluation confidence threshold and the compiled-execution mode. The four inference-only configurations (L0, L1, L2, L4) were scored at the CLI's default confidence threshold of $0.01$, whereas the L3 fine-tuning run's built-in evaluation pass was launched with an explicit threshold of $0.05$. Because COCO AP is rank-based under a per-image detection cap, a higher threshold discards low-confidence detections that could still have contributed to the precision--recall tail, so this difference is expected to work \emph{against} L3 rather than for it. We cannot quantify the size of the effect: the evaluation server is closed, so L3 cannot be re-scored at $0.01$. We therefore report the L1--L3 comparison as confounded by this setting in addition to the optimization differences already noted, and we do not treat the L1--L3 gap as an isolated measurement of fine-tuning. The remaining inference settings (top-$k$ \texttt{num\_select} $=300$ and coordinate restoration) were the shared defaults across all five runs; L0 additionally ran with the compiled-inference path enabled, which the resolution overrides in L1, L2 and L4 necessarily disable. Accordingly, the L0--L1 comparison---our headline result---is weight-matched but not execution-path-matched. The compiled and eager paths are intended to implement the same detector, but their numerical equivalence was not separately verified on the closed benchmark. The verbatim platform commands for all five runs are listed in the supplementary material.

\section{Evaluated Configurations}

\begin{figure}[!ht]
\centering
\resizebox{\textwidth}{!}{%
\begin{tikzpicture}[
  font=\small,
  node distance=6mm,
  box/.style={draw, rounded corners=2pt, align=center, inner sep=3pt, minimum height=7mm},
  lbl/.style={font=\scriptsize\itshape},
]
\node[box, fill=black!5] (ckpt) {704-trained checkpoint};
\node[box, below left=8mm and 28mm of ckpt] (frozen) {parameters \textbf{frozen}};
\node[box, below right=8mm and 28mm of ckpt] (upd) {parameters \textbf{updated}};
\node[box, below=9mm of frozen, xshift=-41mm] (l0) {L0\\$704^2$\\{\scriptsize$\tau{=}.01$, compiled}};
\node[box, below=9mm of frozen, xshift=-14mm] (l1) {\textbf{L1}\\$1120^2$\\{\scriptsize$\tau{=}.01$, eager}};
\node[box, below=9mm of frozen, xshift=13mm] (l2) {L2\\$1280{\times}736$\\{\scriptsize$\tau{=}.01$, eager}};
\node[box, below=9mm of frozen, xshift=41mm] (l4) {L4\\$1120^2{+}$GW\\{\scriptsize$\tau{=}.01$, eager}};
\node[box, below=9mm of upd] (l3) {L3\\$1120^2$ FT\\{\scriptsize\bfseries$\tau{=}.05$, eager}};
\draw (ckpt) -- (frozen); \draw (ckpt) -- (upd);
\draw (frozen) -- (l0); \draw (frozen) -- (l1); \draw (frozen) -- (l2); \draw (frozen) -- (l4);
\draw (upd) -- (l3);
\node[lbl, below=2mm of l1, xshift=6mm] {inference-time processing only};
\node[lbl, below=2mm of l3] {warm-start fine-tuning};
\end{tikzpicture}}
\caption{Overview of the archived comparison. L0, L1, L2 and L4 share the same 704-trained checkpoint and apply no parameter update; L3 warm-starts from it and updates the parameters. Each run's confidence threshold $\tau$ and execution mode are shown beneath it---the two settings the run-record audit found non-uniform (Section~\ref{sec:threshold}). Every configuration was evaluated once on aggregate hidden-server metrics, without domain-separated feedback.}
\label{fig:overview}
\end{figure}
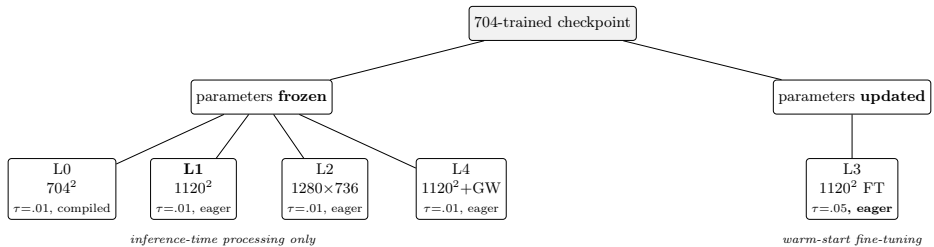

We analyze five archived configurations, holding the backbone fixed unless stated: three intended interventions, the 704 baseline, and one rectangular run that the audit showed to have used an unintended orientation. L0, L1, L2, and L4 are inference-only on one shared checkpoint; L3 updates parameters. Figure~\ref{fig:overview} summarizes the comparison.

\begin{description}
\item[L1 --- Frozen 1120 inference.] This training-free intervention raises the input resolution from $704\times704$ to a square $1120\times1120$ via direct resize, with positional embeddings interpolated by the RF-DETR implementation, while keeping the model parameters frozen. In this implementation the resolution override also forces eager execution instead of the compiled native-resolution path L0 uses (Section~\ref{sec:threshold}); no parameter update was performed. No gradient update, test-time optimization, or pseudo-labeling. Confidence threshold, top-k (num\_select 300), and coordinate restoration follow the base inference config; the exact confidence threshold, resize/interpolation, coordinate-restoration, and box-clipping settings are given in the supplementary configuration.
\item[L2 --- Rectangular inference.] Square $1120\times1120$ vs.\ a rectangular $1280(H)\times736(W)$ resize on the same frozen checkpoint. The benchmark CLI parses rectangular resolutions in $H\times W$ order, so the recorded string \texttt{1280\allowbreak x736} denotes a \emph{tall} input, not a width-major one. The two inputs differ in orientation, aspect ratio, image shape, \emph{and} total pixel count (1.25M vs.\ 0.94M), so this comparison isolates none of these factors and should not be read as aspect-ratio-preserving or native-aspect inference.
\item[L3 --- 1120px fine-tuning (warm-start).] Warm-start from the shared 704px checkpoint (the same checkpoint used for all inference-only configurations, L0--L2 and L4) and fine-tune at $1120\times1120$ for 6 additional epochs with AdamW, a cosine schedule, and mixed precision, at effective batch 16 (batch 1 $\times$ grad-accum 16); the optimizer and scheduler are re-initialized, with encoder and decoder learning rates of $1\times10^{-4}$ and $1.5\times10^{-5}$ respectively. We evaluate the fine-tuned checkpoint---selected as the higher-scoring of the EMA and regular weights by source-validation AP---at $1120\times1120$. Its evaluation pass ran at confidence threshold $0.05$ rather than the $0.01$ used by the other runs (Section~\ref{sec:threshold}). The exact warmup for this run was not separately retained; the remaining settings are documented in the supplementary material.
\item[L4 --- Gray-world white balancing.] Per-image gray-world channel-mean normalization at inference on the frozen checkpoint.
\end{description}

\FloatBarrier
\section{Main Results}

All numbers are aggregate AP\,/\,AR on the \textbf{full} hidden benchmark. The base model is a single RF-DETR-Large trained at 704px; each configuration implements the indicated intervention, with two run-level settings not uniform across the five runs (Section~\ref{sec:threshold}). AP is averaged over IoU $0.50{:}0.05{:}0.95$; best value per column in bold. $\Delta$AP is the change in aggregate AP relative to the 704 base (L0).

\begin{table}[!ht]
\centering
\caption{Detection precision on the full aggregate hidden benchmark. All configurations were submitted once; AP is COCO AP averaged over IoU thresholds $0.50{:}0.05{:}0.95$.}
\label{tab:prec}
\setlength{\tabcolsep}{4.5pt}
\resizebox{\textwidth}{!}{%
\begin{tabular}{lccccccc}
\toprule
Configuration & AP & $\Delta$AP & AP$_{50}$ & AP$_{75}$ & AP$_S$ & AP$_M$ & AP$_L$ \\
\midrule
L0 $\cdot$ Base, 704 inference & 0.3272 & --- & 0.4391 & 0.3335 & 0.0519 & 0.1528 & 0.4687 \\
\textbf{L1 $\cdot$ frozen 1120 (training-free)} & \textbf{0.3654} & \textbf{+0.0382} & \textbf{0.4879} & \textbf{0.3857} & 0.0660 & \textbf{0.1924} & \textbf{0.5034} \\
L4 $\cdot$ frozen 1120 $+$ gray-world & 0.3647 & +0.0375 & 0.4870 & 0.3847 & 0.0658 & 0.1917 & 0.5031 \\
L3 $\cdot$ 1120px fine-tuning & 0.3470 & +0.0198 & 0.4626 & 0.3650 & \textbf{0.0678} & 0.1832 & 0.4811 \\
L2 $\cdot$ rectangular $1280(H)\times736(W)$ & 0.3057 & $-0.0215$ & 0.4419 & 0.3173 & 0.0501 & 0.1447 & 0.4337 \\
\bottomrule
\end{tabular}}
\end{table}

\begin{table}[!ht]
\centering
\caption{Detection recall values retained in the archived server reports for L0, L1, and L3.}
\label{tab:recall}
\setlength{\tabcolsep}{8pt}
\begin{tabular}{lccc}
\toprule
Configuration & AR@1 & AR@10 & AR@100 \\
\midrule
L0 $\cdot$ base 704 & 0.3568 & 0.5782 & 0.6174 \\
L1 $\cdot$ frozen 1120 & 0.3827 & 0.6063 & 0.6520 \\
L3 $\cdot$ 1120px fine-tuning & 0.3220 & 0.5131 & 0.5320 \\
\bottomrule
\end{tabular}
\end{table}

\FloatBarrier

AR was retained only for L0, L1, and L3 in our archived evaluation records; because the evaluation server is now closed, the missing L2/L4 AR values cannot be recovered or recomputed.

\paragraph{Reading the tables.} Among the evaluated configurations, frozen \textbf{L1 (0.3654)} achieved the highest aggregate AP, \textbf{+0.0382} over the 704 base (L0, 0.3272) with \emph{zero} parameter updates. The relative gain is largest for small objects (AP$_S$ $+27.2\%$) while the largest absolute gain is for medium objects (AP$_M$ $+0.0396$; small $+0.0141$, large $+0.0347$)---consistent with the hypothesis that higher-resolution inference primarily helps objects occupying fewer input pixels. The evaluated 1120px fine-tuning (L3, 0.3470) scored below L1, and rectangular inference (L2, 0.3057) below even the 704 base. The L1 gain is also not confined to permissive IoU matching: AP$_{75}$ rises by $+0.0522$ over L0, slightly more than the AP$_{50}$ increase of $+0.0488$. We report this pattern descriptively; the aggregate server output cannot establish that it reflects improved localization specifically.

\begin{figure}[!ht]
\centering
\begin{tikzpicture}[font=\small]
% ---- left panel: absolute Delta AP ----
\begin{scope}
  \def\sa{75}   % cm per unit AP
  \draw[->] (0,0) -- (0,3.6) node[above,font=\scriptsize,align=center] {absolute $\Delta$AP};
  \foreach \v in {0,0.01,0.02,0.03,0.04}{
    \draw[gray!35] (0,{\v*\sa}) -- (4.3,{\v*\sa});
    \node[left,font=\scriptsize,gray!70] at (0,{\v*\sa}) {\v};
  }
  \draw (0,0) -- (4.3,0);
  \foreach \x/\val/\lab in {0.55/0.0141/{small}, 1.85/0.0396/{medium}, 3.15/0.0347/{large}}{
    \fill[black!70] (\x,0) rectangle ({\x+0.75},{\val*\sa});
    \node[above,font=\scriptsize] at ({\x+0.375},{\val*\sa}) {$+\val$};
    \node[below,font=\scriptsize] at ({\x+0.375},0) {\lab};
  }
\end{scope}
% ---- right panel: relative gain ----
\begin{scope}[xshift=6.1cm]
  \def\sr{0.115} % cm per percentage point
  \draw[->] (0,0) -- (0,3.6) node[above,font=\scriptsize,align=center] {relative gain (\%)};
  \foreach \v in {0,10,20,30}{
    \draw[gray!35] (0,{\v*\sr}) -- (4.3,{\v*\sr});
    \node[left,font=\scriptsize,gray!70] at (0,{\v*\sr}) {\v};
  }
  \draw (0,0) -- (4.3,0);
  \foreach \x/\val/\lab in {0.55/27.2/{small}, 1.85/25.9/{medium}, 3.15/7.4/{large}}{
    \fill[black!40] (\x,0) rectangle ({\x+0.75},{\val*\sr});
    \node[above,font=\scriptsize] at ({\x+0.375},{\val*\sr}) {$+\val\%$};
    \node[below,font=\scriptsize] at ({\x+0.375},0) {\lab};
  }
\end{scope}
\end{tikzpicture}
\caption{Effect of frozen 1120 inference (L1) over the 704 base (L0), by COCO object-size class. \textbf{The two orderings differ}, so both are plotted on their own axes rather than one being encoded as a label: medium objects show the largest \emph{absolute} gain (left, $+0.0396$) while small objects show the largest \emph{relative} gain (right, $+27.2\%$). This descriptive scale pattern is compatible with, but does not establish, a small-object resolution mechanism. Single submission per configuration; no error bars are available.}
\label{fig:scale}
\end{figure}
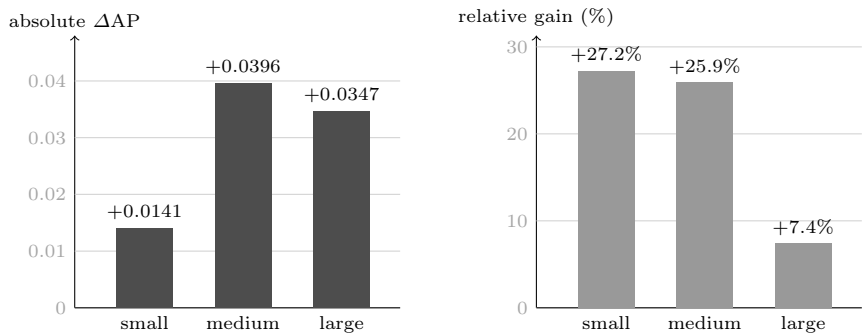

\paragraph{How much of this is signal?} No configuration was repeated and no seed was fixed, so we have no direct estimate of run-to-run variance; the following reading is qualitative, and the three gaps in Table~\ref{tab:prec} rest on three \emph{different} kinds of argument that we keep separate.
\begin{itemize}
\item \textbf{L1 vs.\ L4 ($0.0007$ AP).} Two \emph{different} interventions land within a hair of each other. We do \textbf{not} treat this near-tie as a variance estimate---it compares gray-world against its absence, not a run against a replicate of itself---but it gives us no basis to order L1 above L4, so we report their ordering as undetermined.
\item \textbf{L1 vs.\ L3 ($0.0184$ AP).} This is not a question of noise at all: L1 and L3 differ in the four recipe components and one unplanned evaluation mismatch of Table~\ref{tab:l1l3}, so the gap does not isolate any of them regardless of its magnitude.
\item \textbf{L0 vs.\ L1 ($0.0382$ AP).} The largest gap, and the only comparison free of any optimization difference---its one nuisance variable, compiled-versus-eager execution, is discussed in Section~\ref{sec:threshold}. We treat it as our most robust observation while stressing that, absent repeated runs, we attach no confidence interval to it.
\end{itemize}
Separately, the best score was selected after comparing several hidden-server submissions, which introduces evaluation-set selection bias in the usual winner's-curse sense. A small set of same-configuration replicates on a local source-validation split would convert this qualitative ordering into a quantitative one, but the platform's data-access rules and the closed server prevented it; we flag it as the cleanest available strengthening of the L0--L1 claim.

\section{Analysis}

\subsection{Interpreting the archived fine-tuning outcome}

Table~\ref{tab:l1l3} lists the joint differences between the complete L1 and L3 recipes. The evaluation-threshold mismatch was not part of the intended comparison.

\begin{table}[!ht]
\centering
\caption{Joint differences between frozen 1120 inference (L1) and the evaluated 1120px fine-tuning recipe (L3). The optimization, training-resolution and checkpoint-selection differences are \emph{components of the L3 recipe}; the confidence-threshold difference is an unplanned evaluation mismatch. The comparison therefore evaluates the complete archived recipes rather than isolating parameter updates. Scores: L1 AP $0.3654$, L3 AP $0.3470$ (full breakdown in Table~\ref{tab:prec}).}
\label{tab:l1l3}
\setlength{\tabcolsep}{4pt}
\resizebox{\textwidth}{!}{%
\begin{tabular}{lccl}
\toprule
Factor & L1 (frozen) & L3 (fine-tuned) & Interpretation \\
\midrule
Parameter update & none & 6 epochs & part of the evaluated L3 recipe \\
Training resolution & 704 (inference 1120) & 1120 & part of the evaluated L3 recipe \\
Optimizer / scheduler & --- & re-initialized & part of the evaluated L3 recipe \\
Checkpoint selection & shared 704 ckpt & best of EMA/regular @1120 & part of the evaluated L3 recipe \\
Evaluation threshold & $0.01$ & $0.05$ & \textbf{unplanned} evaluation mismatch \\
\bottomrule
\end{tabular}}
\end{table}

1120px warm-start fine-tuning (L3) raised in-domain validation AP (base 0.767 $\rightarrow$ fine-tuned 0.789) while reaching a \emph{lower} aggregate hidden-benchmark AP than frozen 1120 inference ($0.3654 \rightarrow 0.3470$). Two details cut against reading this as a clean fine-tuning failure. First, the recall drop (AR@100 $0.6520 \rightarrow 0.5320$) is the metric most exposed to the confidence threshold: raising the threshold from $0.01$ to $0.05$ mechanically reduces the set of retained low-confidence detections and can reduce attainable recall. The AR gap is therefore likely affected by L3's higher evaluation threshold and cannot be read as evidence that fine-tuning harmed recall; without re-scoring L3 at $0.01$, the relative contributions of thresholding and training-side changes remain unknown. Second, L3 attains the \emph{highest} small-object AP among the archived configurations (AP$_S$ $0.0678$ vs.\ L1's $0.0660$) despite using a higher threshold that may suppress low-confidence detections---an observation that does not support a claim of uniform degradation from the evaluated recipe. What remains is a modest aggregate-AP difference compatible with several explanations---specialization to the labeled source distribution, confidence calibration, threshold truncation, checkpoint selection, optimizer/scheduler effects---which the aggregate server output cannot separate. We list them as parallel candidates and endorse none. We emphasize that L1 and L3 differ in more than resolution (optimizer updates, epochs, schedule, checkpoint selection, and the evaluation confidence threshold of Section~\ref{sec:threshold}), so we conclude only that, under the specific warm-start fine-tuning recipe \emph{and evaluation settings} used, frozen 1120px inference achieved a higher aggregate score.

\FloatBarrier
\subsection{Negative results, precisely}
\begin{itemize}
\item \textbf{Rectangular input (L2; unintended portrait orientation).} This run was intended as a landscape, roughly native-aspect input, matching the predominantly $1920\times1080$ landscape imagery of the benchmark. The CLI parses rectangular resolutions in $H\times W$ order, so the string \texttt{1280x736} instead produced a \emph{portrait} $1280$-high $\times$ $736$-wide input. We therefore report it as a run-audit finding rather than as evidence about the intended landscape or native-aspect configuration. Rectangular $1280(H)\times736(W)$ inference scored 0.3057---below the 704 base (0.3272). Its AP$_{50}$ (0.4419) is comparable to the base (0.4391) while AP$_{75}$ is lower (0.3173 vs.\ 0.3335), suggesting, but not establishing, that the degradation may be more closely related to localization precision than to coarse object discovery. Because orientation, aspect ratio, image shape, and pixel count all change together, this comparison does not isolate any single factor; we attribute the L1 gain to \emph{square} high-resolution inference rather than to input geometry alone.
\item \textbf{Gray-world (L4).} Inference-time gray-world normalization produced a negligible difference of $-0.0007$ AP ($0.3654 \rightarrow 0.3647$); with no repeated runs we report the measured difference but do not interpret it as evidence that gray-world normalization is detrimental.
\end{itemize}

\section{Limitations and validity}
\label{sec:limits}
We collect here the constraints that bound every claim in the paper; earlier sections refer back to this section rather than restating them.
Table~\ref{tab:boundary} pairs each claim with the evidence it rests on and the scope outside which we do not assert it; the aggregate-only metric, the unmatched L1--L3 control and the single-model scope are recorded there as claim boundaries rather than repeated here. Three constraints are not specific to any single claim.

\begin{itemize}
\item \textbf{Single run ($n{=}1$).} Each configuration was submitted once and no fixed random seed was set; we report no error bars, significance tests, or prediction-level paired statistics. The inference-only comparisons are \emph{checkpoint-shared}, not statistically paired, and not execution-path-matched (Section~\ref{sec:threshold}).
\item \textbf{Model selection on the hidden server.} We compared a limited set of configurations and kept the best. We did not run repeated leaderboard-driven hyperparameter tuning, but selecting among several submitted configurations still introduces evaluation-set selection bias; the best reported result is an exploratory challenge result, not an unbiased estimate on an untouched test set. For scale, our best entry scored $0.3654$ while the leading Track 6 entries scored $\approx0.475$---about $0.11$ AP above ours, roughly three times the largest effect reported in this paper.
\item \textbf{Inference cost and unrun analyses.} Frozen 1120 inference processes $2.53\times$ the pixels of 704; we make no efficiency claim, and we did not measure latency or VRAM because after the challenge the compute budget was exhausted and the evaluation server was closed. The same closure, together with the fact that benchmark images, ground-truth annotations and per-image outcomes were never exposed to participants, prevented a resolution sweep, a matched same-epoch/same-threshold 704 fine-tuning control, class-wise AP, a confidence-calibration study, and any qualitative analysis on the hidden benchmark; these are future work. In particular we make \emph{no} claim that 1120 is an optimal or sufficient resolution: it is the one square high-resolution setting we were able to evaluate, the point of diminishing returns was never located, and intermediate settings (e.g.\ 896) were never submitted.
\end{itemize}

\begin{table}[!ht]
\centering
\caption{Claim, supporting evidence, and the boundary of each claim.}
\label{tab:boundary}
\setlength{\tabcolsep}{4pt}
\resizebox{\textwidth}{!}{%
\begin{tabular}{>{\raggedright\arraybackslash}p{0.30\textwidth}>{\raggedright\arraybackslash}p{0.30\textwidth}>{\raggedright\arraybackslash}p{0.30\textwidth}}
\toprule
Claim & Supporting evidence & Boundary \\
\midrule
Frozen 1120 inference achieved the best tested aggregate AP & L1 AP 0.3654 vs.\ L0 0.3272 & Evaluated configurations only; one model on one benchmark; not a state-of-the-art challenge entry \\
Gains were strongest on small and medium objects & AP$_S$ $+27.2\%$ relative; AP$_M$ $+0.0396$ absolute & Aggregate hidden mixture; no per-domain split \\
The archived L3 run produced a lower aggregate score than L1 & L3 AP 0.3470; AR@100 0.5320 & One recipe; four recipe components plus an unplanned threshold mismatch ($0.05$ vs.\ $0.01$) differ (Table~\ref{tab:l1l3}); the threshold effect is unmeasured and unrecoverable \\
In-domain validation is an unreliable selection signal under domain shift & L3 val $0.767\rightarrow0.789$ while its aggregate AP did not rise & One model and recipe; benchmark-level; suggestive, not proven \\
Rectangular and gray-world variants did not improve on L1 & L2 AP 0.3057 (unintended orientation); L4 AP 0.3647 & One server evaluation each; $n{=}1$ \\
Target-city improvement is \emph{not} established & No domain-separated server metric exists & Benchmark-level interpretation only \\
\bottomrule
\end{tabular}}
\end{table}

\FloatBarrier

\section{Discussion}

Two takeaways survive the constraints of Section~\ref{sec:limits}, and we state them at the level the data support.

First, as a \emph{practical default under restricted feedback}, frozen resolution scaling is worth evaluating before additional optimization: it leaves the trained weights untouched, permits a checkpoint-shared comparison, and here produced the largest gap we observed (L0--L1)---at a higher inference cost ($2.53\times$ the pixels). Second, and more transferable, our one fine-tuning run is a \emph{caution about model selection}: its in-domain validation AP rose ($0.767 \rightarrow 0.789$) while its aggregate benchmark AP did not, so under aggregate-only cross-city feedback in-domain validation ranked a change that the hidden benchmark did not reward.

These findings \textbf{should not} be read as evidence that higher-resolution inference generally beats fine-tuning or domain adaptation. For one RF-DETR-Large model and the specific, confounded configurations evaluated here, frozen inference simply scored highest, and the fine-tuning comparison does not isolate any single factor (Table~\ref{tab:l1l3}).

From the second takeaway we \emph{suggest}, but do not demonstrate, a decision rule for similarly restricted settings: when no target-validation signal is available, prefer a \textbf{domain-shift proxy} validation split---held-out cameras, locations, or acquisition conditions drawn from the source data---over in-domain validation when deciding whether to apply training-side changes, and treat any single-submission leaderboard result as an upward-biased estimate. Building such a proxy split and testing whether it correlates with the hidden benchmark better than in-domain validation is the natural next step; the platform's data-access rules prevented us from doing so here. Future work should also prioritize a matched same-epoch, same-threshold 704px fine-tuning control, a denser resolution--compute sweep, and direct latency/VRAM profiling.

\section{Conclusion}

We presented an audited, reproducible study of a few inference- and training-time configurations of a single frozen RF-DETR-Large detector in AI City Challenge 2026 Track 6, under an aggregate-only cross-city metric and a limited submission budget. Among the configurations we could submit, frozen square-1120px inference achieved the highest aggregate hidden-benchmark AP with no parameter update---the largest relative gain on small objects and the largest absolute gain on medium objects. The one warm-start fine-tuning run raised in-domain validation but not aggregate AP, cautioning that in-domain validation is an unreliable selection signal under domain shift. The rectangular run used an unintended orientation rather than a designed ablation, and gray-world normalization was negligible. We therefore frame this paper not as a competitive system but as a carefully qualified challenge study: a run-record audit, an archived and confounded warm-start fine-tuning outcome under non-uniform evaluation settings, and honest negative interventions under aggregate-only cross-city evaluation---not a general law. Code, configuration files, and inference-packaging scripts are included in the supplementary material and are publicly available at \url{https://github.com/freak-jaeuk/aicity2026-track6}, commit \texttt{763021c}. Model weights and the challenge datasets are not redistributable and are therefore not included.

\bibliographystyle{splncs04}
\bibliography{refs}

\end{document}